\documentclass[conference]{IEEEtran}
\IEEEoverridecommandlockouts
\usepackage{cite}
\usepackage{amsmath,amssymb,amsfonts}
\usepackage{algorithmic}
\usepackage{graphicx}
\usepackage{textcomp}
\usepackage[caption=false]{subfig}
\usepackage{xcolor}
\usepackage{url}

\usepackage{fancyhdr}
\begin{document}

\title{AffectGAN: Affect-Based Generative Art Driven by Semantics

\thanks{This project has received funding from the European Union’s Horizon 2020 programme under grant agreement No 951911.}
}

\author{Anonymous}
\author{
\IEEEauthorblockN{Theodoros Galanos}
\IEEEauthorblockA{\textit{Institute of Digital Games} \\
\textit{University of Malta}\\
Msida, Malta \\
theodoros.galanos.21@um.edu.mt}
\and
\IEEEauthorblockN{Antonios Liapis}
\IEEEauthorblockA{\textit{Institute of Digital Games} \\
\textit{University of Malta}\\
Msida, Malta \\
antonios.liapis@um.edu.mt}
\and
\IEEEauthorblockN{Georgios N. Yannakakis}
\IEEEauthorblockA{\textit{Institute of Digital Games} \\
\textit{University of Malta}\\
Msida, Malta \\
georgios.yannakakis@um.edu.mt}
}

\maketitle
\thispagestyle{fancy}

\begin{abstract}
This paper introduces a novel method for generating artistic images that express particular affective states. Leveraging state-of-the-art deep learning methods for visual generation (through generative adversarial networks), semantic models from OpenAI, and the annotated dataset of the visual art encyclopedia WikiArt, our \emph{AffectGAN} model is able to generate images based on specific or broad semantic prompts and intended affective outcomes. A small dataset of 32 images generated by AffectGAN is annotated by 50 participants in terms of the particular emotion they elicit, as well as their quality and novelty. Results show that for most instances the intended emotion used as a prompt for image generation matches the participants' responses. This small-scale study brings forth a new vision towards blending affective computing with computational creativity, enabling generative systems with intentionality in terms of the emotions they wish their output to elicit.
\end{abstract}


\section{Introduction}\label{sec:introduction}

\begin{figure}
\centering
\hfill
\subfloat[``A happy cityscape'']{\includegraphics[width=0.48\columnwidth]{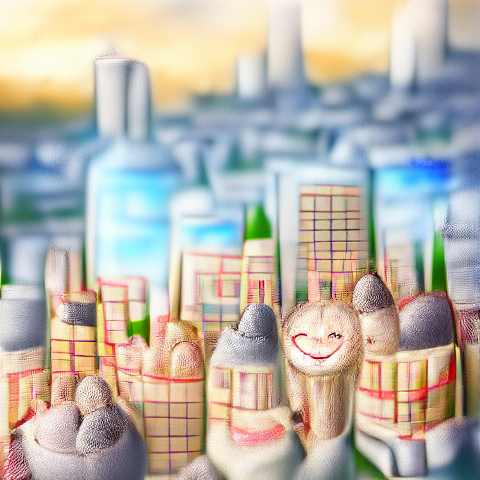}\label{fig:best1}}\hfill
\subfloat[``A depressed cityscape'']{\includegraphics[width=0.48\columnwidth]{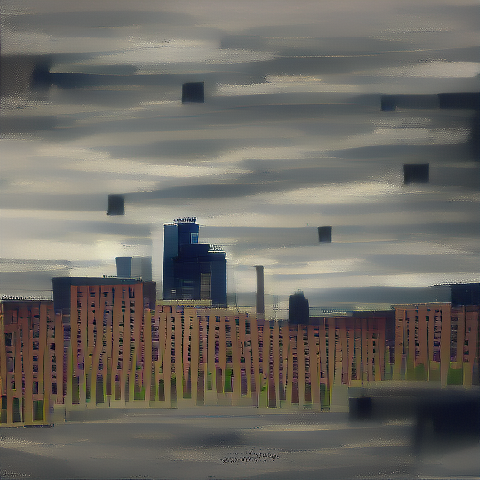}\label{fig:best2}}
\hfill
\caption{Two indicative images generated via AffectGAN for a cityscape prompt.}
\label{fig:best}
\end{figure}

Affective computing is predominately relying on affective datasets containing elicitors that come in the form of sounds, videos, games, and images. Those elicitors are first annotated with affect labels and are then processed by affect models to derive a mapping between the context of the elicitor (e.g. the frames of a video, or the pixels of an image) and its corresponding label. The expression of affect that builds on these models usually takes a direct human-centric form such an avatar's facial expression, a virtual human's body stance, or the tone of synthesised speech. In this work, instead, we build on state of the art machine learned representations and propose a new method, namely \emph{AffectGAN}, for the semantic generation of images that embed affective information and thereby, express such information via art (see Fig \ref{fig:best}). The paper acts as a bridge between the fields of \emph{affective computing} and \emph{computational creativity} \cite{colton2009coming} for the benefit of both: the former is equipped with visual art-based affect expression capacities; the latter benefits from the affect-driven generation capacities of the proposed method.

To reliably generate affect-infused images via semantics, AffectGAN integrates two models: OpenAI's Contrastive Language–Image Pre-training (CLIP) \cite{radford2021learning} and a convolutional VQGAN \cite{esser2020taming}. The CLIP model enables the embedding of semantic information to the generative process as it is a multimodal model trained on image and text pairs. Therefore, a trained CLIP model has learned to associate visual and language information. The VQGAN model is trained on the publicly available WikiArt dataset \cite{mohammad2018wikiart}. VQGAN extracts and learns a visual vocabulary from training, called a \emph{codebook}, on top of a quantized representation of the image (i.e. patch-based); the image composition is modeled with an autoregressive transformer. The combination of CLIP and VQGAN brings together the benefits of expressing visual information through semantics (CLIP) and conditionally synthesising an image through the conditions imposed by semantics (VQGAN). The generation process of AffectGAN follows iterative optimization, with VQGAN providing an image in the form of 16x16 patches and CLIP providing the reward signal in optimizing each patch according to the given semantic prompt.

To evaluate AffectGAN we pick four indicative affective states placed at each of the four quadrants of the arousal-valence circumplex model \cite{circumplex2018}: \textit{depression}, \textit{calmness}, \textit{anger}, and \textit{happiness}. We then let AffectGAN generate four images across eight typical painting types (e.g. landscapes or portraits) for the four target emotions resulting in 32 images. The images were generated using the following semantic prompt: ``A $\langle$\emph{emotion}$\rangle$ $\langle$\emph{genre}$\rangle$''---e.g. ``a happy citycape'' in Fig.~\ref{fig:best}. An online user study with 50 participants evaluated the emotions elicited per image, as well as its quality and novelty \cite{ritchie2007criteria}. Our findings suggest that participants were most successful at matching the emotion expressed by AffectGAN in depression prompts (68\%), followed by anger (65\%), calmness (62\%) and happiness (36\%). A further analysis of results shows that both the affect prompt and the semantic prompt (i.e. painting type) influenced users' feedback in terms of emotions elicited as well as quality and novelty valuations.

\section{Background}\label{sec:background}

The relationship between affective expression and computational art is well-studied. Within the field of computational creativity a number of studies have investigated how music \cite{scirea2015moody,monteith2010music} or images \cite{colton2015sees} can be generated to represent certain concepts. In computational creativity research, endowing an algorithmic system with the ability to possess some intent and be able to express it is paramount \cite{colton2011face}. In many cases, this intent is achieved by communicating the semantic concept that underlies the created artefacts. DARCI \cite{ventura2019autonomous} is such an image generation system that couples neural networks and genetic algorithms to create images that correspond to adjectives. In its early version \cite{heath2016creating} Heath and Ventura trained vector representations of words that could describe a given image; these vectors could guide the process of rendering images that convey particular concepts. That line of work is similar to ours but the focus and scope are different: AffetGAN puts an emphasis on affect-infused image generation and its input can contain any semantic prompt provided by the artist. Importantly, while DARCI has had several versions and implementations, its main method of generating artefacts is by automatically programming image processing filters for an original input image. Instead, AffectGAN does not rely on any human-provided ``seed'' image and synthesizes the visual results on its own.

In affective computing, images or frames of videos of humans are predominately used to model user affect through its manifestations on facial expressions and body stance \cite{calvo2010affect}. A recent set of studies in affect modeling \cite{makantasis2019pixels,makantasis2021pixels}, however,  reframes that perspective and instead takes a user-agnostic approach by training arousal models solely on pixels and sounds of the user interaction. We follow the latter path in this work and build a generative system that expresses affect that does not rely on images of users. Instead we rely on affect labels provided on images---via the WikiArt dataset \cite{Tan2016CeciNP}---and semantic information as provided by the users of AffectGAN.   

In the last few years, contrastive learning approaches \cite{chen2020simple,he2020momentum} have transformed Computer Vision and opened the door to many applications that were only recently  considered impossible. OpenAI's CLIP model \cite{radford2021learning} was the first to truly show the potential of a relatively simple contrastive learning architecture when scaled to large datasets (in the order of 400 million images) and training computation. Before these datasets and computing power was available, earlier works experimented with generating images using text, using both transformers \cite{mansimov2015generating} and Generative Adversarial Networks (GANs) \cite{reed2016generative,xu2018attngan,stackgan,li2019object}. However, the quality of images generated by such approaches never showed satisfactory results for complex domains. The recent surge of multimodal models trained on newly developed multimodal datasets containing both images and semantic annotations, along with novel (mostly) transformer-based architectures for both text and image data modalities, has suddenly made high quality image generation through language feasible.


\section{AffectGAN}\label{sec:methodology}

The process of generating images based on affect has two main components. The first model, OpenAI's CLIP \cite{radford2021learning}, is used to enable the injection of semantic information to the generation process. CLIP is a mulimodal contrastive learning model that was trained on 400 million pairs of images and text annotations collected by OpenAI. There are a few variants of pretrained models that have been made public, with the most popular being the model based on the Vision Transformer architecture introduced in \cite{dosovitskiy2020image}, specifically the ViT/32 variant. CLIP has been incredibly successful in enabling a new generation of AI artists in many creative workflows and generative processes and is perhaps one of the most important pretrained models ever released. The second component is a VQGAN model \cite{esser2020taming}, a state of the art generative model that follows a hybrid architecture with both transformer and GAN elements. A major advantage of the VQGAN model is its ability to learn quantized embeddings of images (specifically image patches), a visual vocabulary that is highly efficient in generating high quality images. For the purpose of AffectGAN, a custom VQGAN model was trained on the WikiArt dataset (first released in \cite{Tan2016CeciNP}), a dataset of $81,444$ images of artistic creations (paintings, images) across many different art styles. The model was trained for 7 million steps, where in each step the model `sees' one random crop from an image in the training set. The model was then fine-tuned on a higher resolution version of the dataset\footnote{https://archive.org/details/wikiart-dataset} for an additional 1 million steps. This improved the visual quality of reconstructions, since it provided access to larger resolution crops to the model, which had more visual content, as well as help remove certain JPEG artifacts that were evident after the first training on the smaller resolution dataset. 

\begin{figure}[!tb]
\centering
\includegraphics[width=\columnwidth]{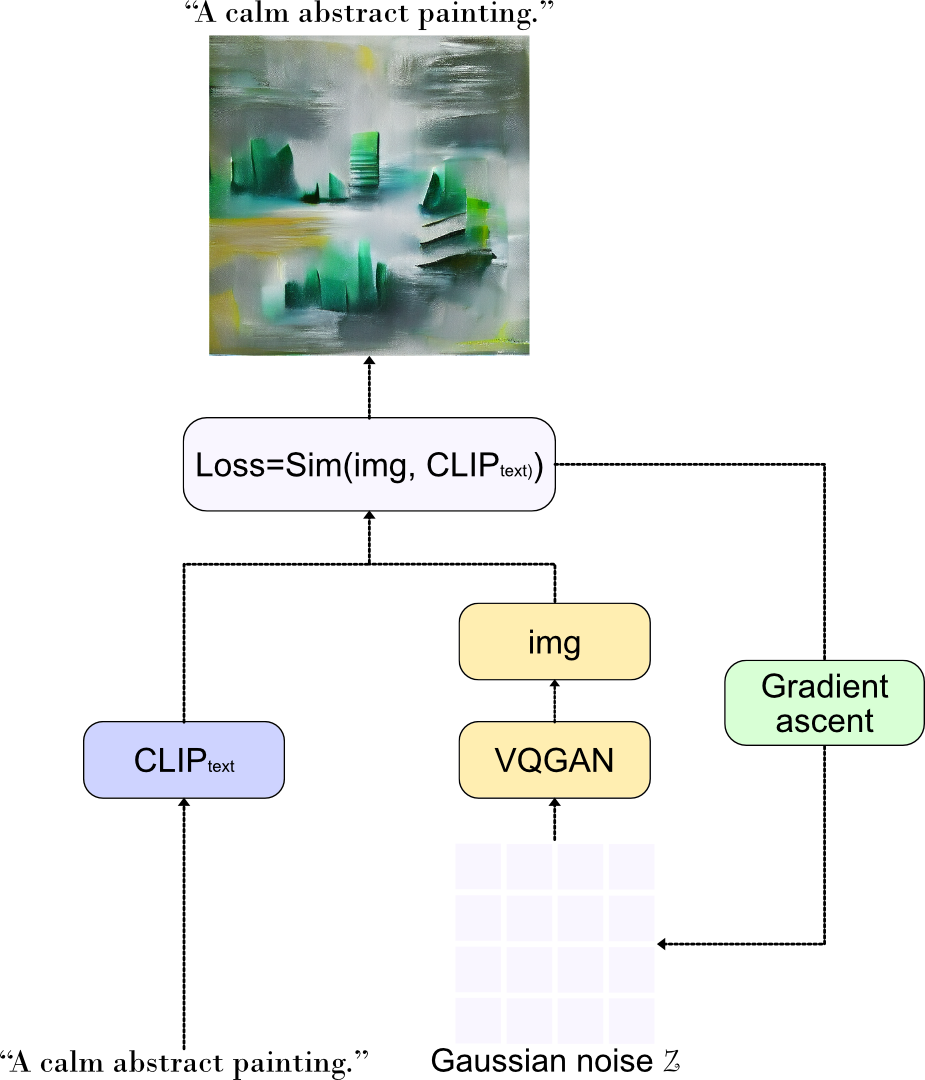}
\caption{Workflow of the image generation process}\label{fig:workflow}
\end{figure}

The affective generation process combines the two models in an iterative optimization called \textit{codebook sampling}. During this process, we use a grid of independent categorical distributions, parameterized by the logits of each class (or code). Each grid cell is 16x16 pixels, and their logits are initialized with Gaussian noise. The total number of possible codes is defined by the properties of the VQGAN model in use, specifically the size of its codebook. In our case, the WikiArt model used has 1024 codes. The randomly initialized logits pass through a softmax layer to get the probability of each code in the codebook and then sample from each distribution independently. Finally, we feed these samples to the VQGAN model and generate an image. 
We minimize the CLIP loss (cosine similarity) between the output image and the text embedding of the prompt used to generate the image via the AdamW optimizer. To generate the output image that the user eventually sees, we take the argmax of the final logits in order to select the respective VQGAN visual codes, instead of sampling. 

The combination of CLIP and VQGAN, and others like it (e.g. \cite{galatolo2021generating,bau2021paint}) has been able to create a vast array of artistic creations in the last few months, opening the doors to semantically guided generative processes in all creative domains.

\section{Results}\label{sec:results}

In this section we outline the experimental results obtained with AffectGAN. In particular, we describe the experimental protocol we designed for validating the tool and go through the details of our participants. Finally, we discuss the key findings obtained through our survey.

\subsection{Experimental Protocol}

In order to assess the efficacy of the semantic generation process in creating images based on affective descriptions, we generated a set of images with different genre and emotional context and conducted a user questionnaire on their qualities.

To produce a small but diverse dataset of generated images for user evaluation, four affective prompts were used---one emotion from each quadrant of the two-dimensional arousal-valence space defined by the circumplex model \cite{circumplex2018}. The two low arousal emotions were \textit{depression} (negative valence) and \textit{calmness} (positive valence), while the two high arousal emotions were \textit{anger} (negative valence) and \textit{happiness} (positive valence). 
The semantic content (i.e. prompts) used to generate the images in this research followed the work done by the ArtEmis project \cite{achlioptas2021artemis}. Specifically, we used 8 painting types of the genre classes used by WikiArt: Abstract, Cityscape, Genre Painting, Landscape, Portrait, Religious Painting, Sketch Study, Still Life. With 8 painting types and 4 emotions, the resulting dataset is 32 images, which is then evaluated in the user study.

The questionnaire was created in Google Forms, and did not collect any identifiable information about the participant. The questionnaire included a page outlining the experiment and requesting the consent of participants, followed by three demographics questions. After these steps, the dataset of 32 images was shown to all users in the same order (randomized once at the start of the experiment). One image was shown per page, with three mandatory multiple-choice questions per image; once these questions were answered, the next image was shown. The first question was ``What emotional state does this image elicit?'', with possible answers being the four emotions used as prompts (calmness, anger, depression, happiness, with their order randomized per image) or an ``other'' option that allowed users to write their own comments. The other two questions had 5-item Likert-scale responses, and assessed the quality\footnote{The question for quality was ``How would you rate this image in terms of quality?'', and responses range from ``Very Poor'' to ``Very Good''.} and novelty\footnote{The question for novelty was ``How would you rate this image in terms of novelty?'' and responses range from ``Not novel at all'' to ``Very novel''.} of the image.

Participation was advertised among the authors' network of contacts (including social media posts) following a combination of snowball and convenience sampling. 

\subsection{Participants}

A total of 50 human participants filled out the survey. Most participants were between 25 and 35 years old (60\%), with 20\% being between 35 and 54 years old and 18\% between 18 and 24 years old. 70\% of participants identified as male, 24\% as female, 1 participant as non-binary, and 2 participants preferred not to answer. When asked how participants would rank their experience with generative art, the average response was 2.96 out of 5, with 32\% rating their experience at 4 or 5 (Very Experienced) out of 5 in the Likert-scale provided.

\subsection{Findings}\label{sec:experiment_findings}

The main research question that the user study aims to answer is whether the affective prompts used as inputs for image generation were correctly predicted by the end-users viewing the generated image. Additional avenues of inquiry related to the impact of the painting type on human evaluation, and how the prompts (emotion and semantic context) affect users' perception of the generated images' quality and novelty.

\subsubsection{Classification of elicited emotion}\label{sec:experiment_findings_confusion}
\begin{table}[!tb]
\centering
\caption{Confusion matrix between expressed affect via the generated image (rows) and the labelled affect (columns).}
\label{tab:confusion_matrix}
\begin{tabular}{l|c|c|c|c|c}
& Anger & Calmness & Depression & Happiness & Other \\ 
\hline\hline
Anger & 65\% & 3\% & 10\% & 4\% & 19\% \\ \hline
Calmness & 2\% & 62\% & 15\% & 8\% & 13\% \\ \hline
Depression & 2\% & 11\% & 68\% & 1\% & 18\% \\ \hline
Happiness & 4\% & 24\% & 11\% & 36\% & 25\% \\ 

\end{tabular}
\end{table}
Table \ref{tab:confusion_matrix} shows the confusion matrix per original affective prompt used for image generation. It is obvious that many participants (19\% on average) preferred to provide their own free-text responses to the elicited emotion question. Participants were most successful at matching the emotion in depression prompts (e.g. ``A depressed still life'', see Fig.~\ref{fig:sample_min_novelty}) at 68\% accuracy, with anger prompts and calmness prompts not far behind (65\% and 62\% respectively). Happiness was often hard to guess in the generated images when it was used as a prompt (36\% accuracy), and participants often chose their own interpretation (``other'' at 25\%) or confused it with calmness (24\%). Happiness was rarely chosen as a response in general (12\% of total responses, with the second rarest choice being Anger at 18\%) which likely points to some reluctance among participants in labelling an image as making them happy. Among freeform responses for images generated as ``happy'', most common were ``Anxiety'', ``Disgust'', ``Fear'' (8 responses each) and ``Confusion'' (7 responses). We further discuss the implications of ``happy'' as a prompt and as a response in Section \ref{sec:discussion}. It is worth noting that generally, low valence prompts were correctly classified as low valence (72\%) and similarly for high valence (65\%) compared to a higher confusion along the arousal axis. This means that participants usually could correctly classify positive or negative emotions in the images but were challenged to identify high arousal versus low arousal.

\subsubsection{Impact of affective prompts on user responses}
\begin{table}[!tb]
\centering
\caption{Data collected per affective prompt. Results are averaged from 8 images generated per affective prompt, and novelty and quality values include the 95\% confidence interval.}
\label{tab:data_affect}
\begin{tabular}{l|c|@{~}c@{~}|@{~}c@{~}|@{~}c@{~}}
Affective prompt & accuracy & un. answers & quality & novelty \\
\hline\hline
Anger & 65\% & 12.1 & 3.00$\pm$0.29 & 3.27$\pm$0.30 \\ \hline 
Calmness & 62\% & 9.9 & 3.36$\pm$0.30 & 3.04$\pm$0.32 \\ \hline 
Depression & 68\% & 10.3 & 3.06$\pm$0.30 & 3.11$\pm$0.29 \\ \hline 
Happiness & 36\% & 13.5 & 3.00$\pm$0.29 & 3.37$\pm$0.29 \\
\end{tabular}
\end{table}
Table \ref{tab:data_affect} summarizes the users' responses in terms of emotion, quality, and novelty split across the different affective prompts used to generate the images. How different prompts affect whether participants can correctly classify the generated images is already discussed in Section \ref{sec:experiment_findings_confusion}. It is also not surprising that for Happiness and Anger prompts the number of unique responses provided (including freeform text) is higher since the ratio of ``other'' responses is higher for these sets (see Table \ref{tab:confusion_matrix}). Interestingly, images generated for Calmness were rated with higher quality but lower novelty overall than other sets, although the differences are not significant at 95\% confidence. Perhaps due to the confusing emotions elicited by images generated for Happiness and Anger (indicated by the many unique answers), these sets received overall higher novelty ratings. We triangulate these findings below by processing them by painting type.

\subsubsection{Impact of painting types on user responses}
\begin{table}[t]
\centering
\caption{Data collected per painting type. Results are averaged from 4 images generated per painting type, and novelty and quality values include the 95\% confidence interval.}
\label{tab:data_painting}
\begin{tabular}{l|@{~}c@{~}|@{~}c@{~}|@{~}c@{~}|@{~}c@{~}}
Genre prompt & accuracy & un. answers & quality & novelty \\
\hline\hline
Abstract & 66\% & 12.8 & 2.75$\pm$0.31 & 2.67$\pm$0.31 \\ \hline 
Cityscape & 61\% & 13.5 & 3.22$\pm$0.29 & 3.53$\pm$0.28 \\ \hline 
Genre Painting & 62\% & 10.8 & 2.95$\pm$0.28 & 3.17$\pm$0.30 \\ \hline
Landscape & 54\% & 11.8 & 3.43$\pm$0.28 & 3.36$\pm$0.29 \\ \hline 
Portrait & 52\% & 12.0 & 3.35$\pm$0.31 & 3.59$\pm$0.28 \\ \hline 
Religious Painting & 61\% & 12.3 & 2.75$\pm$0.30 & 3.23$\pm$0.32 \\ \hline 
Sketch Study & 65\% & 7.0 & 3.23$\pm$0.31 & 3.20$\pm$0.31 \\ \hline 
Still Life & 42\% & 11.5 & 3.17$\pm$0.29 & 2.85$\pm$0.31 \\
\end{tabular}
\end{table}
Table \ref{tab:data_painting} summarizes the users' responses in terms of emotion, quality, and novelty split across the different painting types used as semantic prompts to generate the images. It is obvious that different semantic prompts impact users' correct prediction of the affective prompt, with the least accurate set generated for still life (42\% accuracy) and the most accurate being abstract paintings (66\%) and sketch study (65\%). While target emotion was often correctly predicted, images generated for abstract prompts had the lowest quality and the lowest novelty. Landscapes, which were overall not well-predicted (54\% accuracy), were rated with the highest quality and third-highest novelty. Portraits (with 52\% accuracy) had the highest novelty and second-highest quality scores. As will be discussed below, different semantic prompts lead to very different images of varying quality; this is captured by the users' responses. Moreover, it seems that more ``interesting'' images to users (novel, good, or that inspire them to provide their own comments in terms of emotions) were harder to correctly classify.

\begin{figure}
\centering
\subfloat[``A calm landscape'']{
\includegraphics[width=0.3\columnwidth]{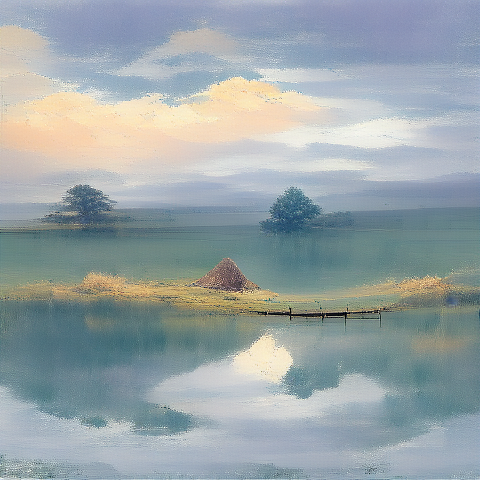}\label{fig:sample_max_accuracy}}~
\subfloat[``A happy sketch and study'']{
\includegraphics[width=0.3\columnwidth]{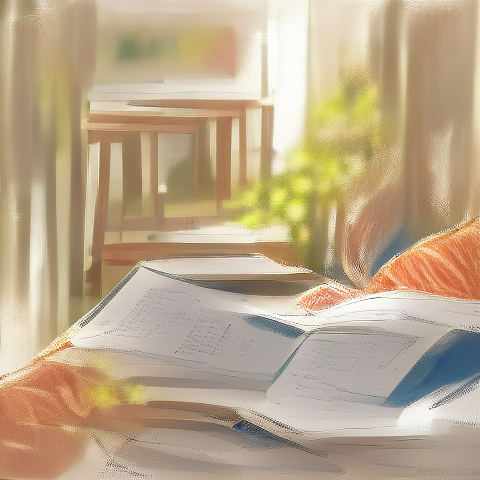}\label{fig:sample_min_accuracy}}~
\subfloat[``An angry landscape'']{
\includegraphics[width=0.3\columnwidth]{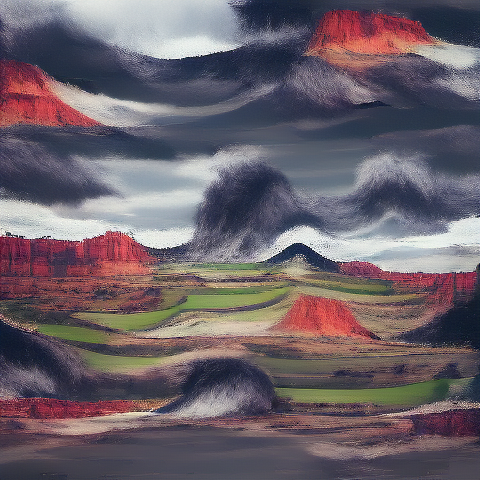}\label{fig:sample_max_answers}}\\
\subfloat[``A happy abstract painting'']{
\includegraphics[width=0.3\columnwidth]{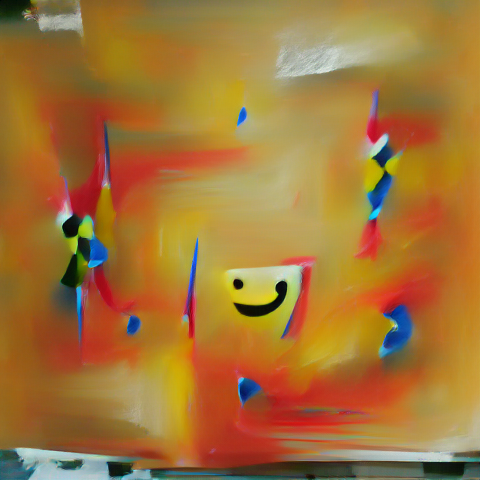}\label{fig:sample_min_quality}}~
\subfloat[``A happy landscape'']{
\includegraphics[width=0.3\columnwidth]{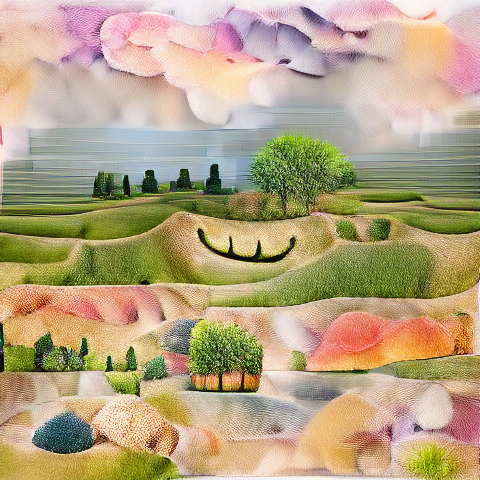}\label{fig:sample_max_novelty}}~
\subfloat[``A depressed still life'']{
\includegraphics[width=0.3\columnwidth]{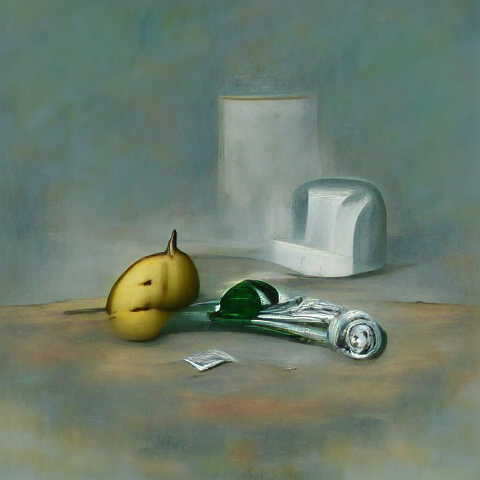}\label{fig:sample_min_novelty}}
\caption{Sample generated images. 
In the user study, Fig.~\ref{fig:sample_max_accuracy} had the highest prediction accuracy (90\%), the fewest unique answers, and the highest quality; Fig.~\ref{fig:sample_min_accuracy} had the lowest accuracy (10\%); Fig.~\ref{fig:sample_max_answers} had the most unique answers;
Fig.~\ref{fig:sample_min_quality} had the lowest quality; Fig.~\ref{fig:sample_max_novelty} had the highest novelty, and Fig.~\ref{fig:sample_min_novelty} had the lowest novelty.
}
\label{fig:samples}
\end{figure}
\subsubsection{Qualitative analysis of generated images}
Figure \ref{fig:samples} shows some indicative images generated by the dataset, chosen as the highest or lowest rated in the different metrics highlighted in Tables \ref{tab:data_affect} and \ref{tab:data_painting}. Some of the images manage to capture fairly realistic settings (e.g. Fig.~\ref{fig:sample_max_accuracy}) or are evocative but not realistic (e.g. Fig.~\ref{fig:sample_max_answers}). There are also glitches with caricature ``faces'' appearing in otherwise unrelated parts of the image (e.g. Fig.~\ref{fig:sample_min_quality}, \ref{fig:sample_min_novelty}), which is discussed in Section \ref{sec:discussion}.

As a preliminary analysis of the patterns in generated images, we quantized each pixel of the image into one of 15 colors (12 colors at equal intervals on the hue scale, as well as white, black and gray) and calculated their ratio to the total image size, similar to \cite{liapis2018pokemon}. At a high level view, images generated for low arousal prompts (Calmness, Depression) have a high ratio of non-saturated pixels (black, white and grey) while those generated for Happiness have far more warm colors (red, orange, or magenta) than other images. Overall, images did not tend to have a high ratio of blue hues (mostly present in Cityscape and Landscape prompts), although images generated for Calmness had more blues than all others. Surprisingly, high ratios of green hues were found in Religious and Genre Paintings, while unsurprisingly the highest ratio of monochrome colors was for Sketch Study prompts. 

The response of users to the color ranges of the images was not as clear-cut as with the original prompts; admittedly, this may be due to the simple metrics used currently for image analysis. That said, significant positive correlations were found between users' ratings of quality and blue hues and between quality and monochrome colors. The former, however, may be due to the fact that Cityscapes and Landscapes (which have more blue colors) were rated higher overall.

\section{Discussion and Next Steps}\label{sec:discussion}

The user study provided a small but informative demonstrator of the power and limitations of AffectGAN. By focusing on different painting types and a small set of diverse emotions (based on their distribution in the arousal-valence axes) we created a compact set of images that had diverse appearance, subject matter, and---admittedly---quality. The users' responses were overall positive, with most generated images (65\%) successfully matched to the affect prompt by at least half of the participants. A weakness identified during the study was the inability of participants to identify happiness in the prompts; this is analyzed further below. Moreover, it appears that images generated for calmness were rated higher in quality but lower in novelty overall than other sets. Different semantic prompts had a impact of the user's affect labeling process with the least accurate set generated for {Still Life} (42\%) and the most accurate---surprisingly---being {Abstract Paintings} (66\%). Images generated for {Abstract Paintings} received the lowest quality and novelty values on average. Overall, results suggest that the hardest images to label correctly in term of affect are the more ``interesting'' images; those are often images of high quality, novelty, or images that inspire annotators to provide their own affective labels.

\begin{figure}
\centering
\subfloat[``An angry religious painting'']{
\includegraphics[width=0.3\columnwidth]{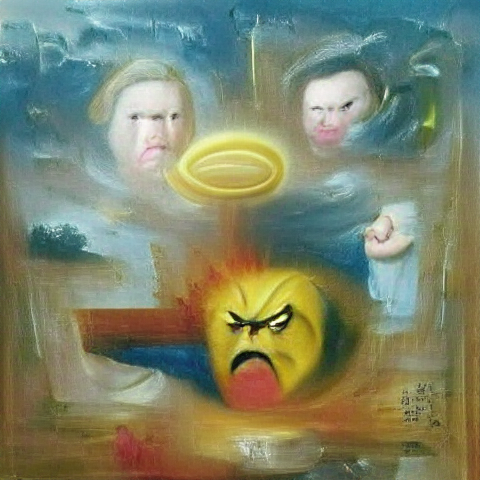}\label{fig:glitch1}
}\hfill
\subfloat[``A depressed landscape'']{
\includegraphics[width=0.3\columnwidth]{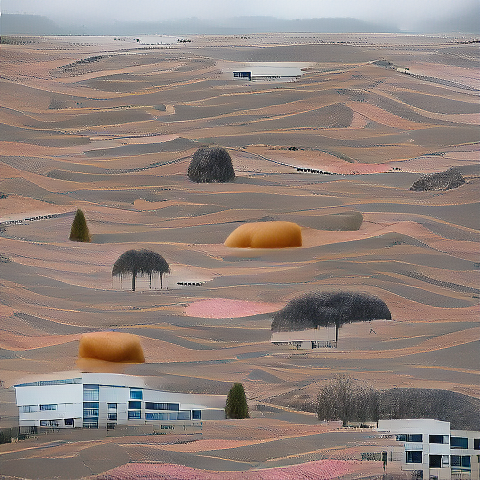}\label{fig:glitch2}
}\hfill
\subfloat[``A happy genre painting'']{
\includegraphics[width=0.3\columnwidth]{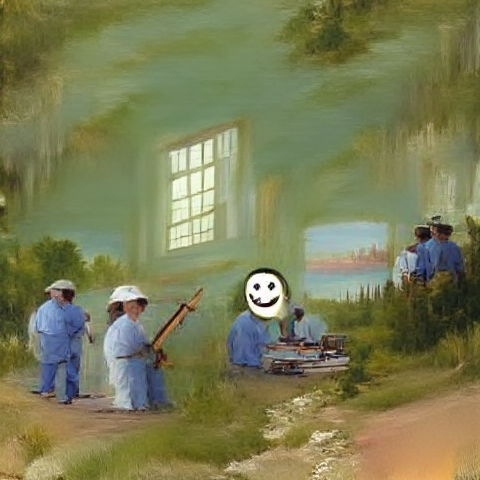}\label{fig:glitch3}
}
\caption{Glitchy images generated via AffectGAN. Fig.~\ref{fig:glitch1} integrates an ``emoji'' face, Fig.~\ref{fig:glitch2} has out-of-place elements, and Fig.~\ref{fig:glitch3} combines the two.}
\label{fig:glitches}
\end{figure}

Generating the 32 images for the user study and processing the participants' responses to them highlighted a number of weaknesses that should be addressed in future iterations of AffectGAN. In terms of the images generated, it was a common occurrence that the image somehow integrated a facial expression indicative of the emotion or, equally often, a stylized ``emoji'' type of glitch. Other glitches more expected from GAN art included out-of-place elements seemingly cut-out and pasted on top of unrelated surrounding art. Examples of both are shown in Fig.~\ref{fig:glitches}. Addressing some of these shortcomings may be challenging, since they are evident in most GAN images generated in the literature. However, the presence of facial expressions and emojis could be tempered by a custom discriminator or post-generation classifier system that detects facial expressions; images that contain faces could be then omitted for specific painting types such as Landscape or Sketch Study. A more pertinent finding from the user study related to generated images for the happiness prompt, which were often mislabeled as calm. There are a number of possible reasons for this. It is our hypothesis that users were more lenient towards associating a static image with feelings of calmness than happiness, since the latter requires more contextual cues. This likely affected both the participants of the user study and the original annotators of the WikiArt dataset, which explains why happiness is often construed with smile emojis as a cognitive, superficial shortcut. Admittedly, none of the other suggested emotions in the high valence and high arousal quadrant are likely to circumvent this weakness. Focusing instead on the other three quadrants with more emotions (e.g. disgust or fear) may be lead to more interesting results in the generated outcomes and in the user responses.

This initial study largely re-used existing code architectures and pre-trained models for a first test case of affect generation. Future work could explore some more tailored inputs, for example by distinguishing emotions as additional inputs rather than as another part of the natural language prompts. A possible direction could e.g. encode a set of emotions in the form of a one-hot encoding, fusing them with the latent vectors produced by CLIP. Another interesting direction is to provide an actual image in conjunction with the affect prompt, allowing GANs to create variants of the same image towards a specific affective outcome. This proposition is similar to the DARCI system \cite{heath2016darci} which generates filters to an existing image based on certain semantic information. Unlike the image processing filters of DARCI, however, AffectGAN can make more substantial changes to the image, introducing new elements etc. Finally, another promising direction for introducing more control to the current implementation of AffectGAN is manipulating the latent vector which represents the final image. This can be achieved through latent vector evolution \cite{deepmasterprints} or by simply combining the latent vectors of two images in some form of interpolation between the two. Beyond the tangible steps for enhancing AffectGAN, more applications that benefit from its representational and generative power should be implemented. Examples include interactive art, where the users can specify the prompts and rate the results, as well as cultural heritage applications where existing images can be manipulated by AffectGAN towards specific evocative dimensions.

\section{Conclusion}\label{sec:conclusion}

AffectGAN is introduced as a generative system of two modules (CLIP and VQGAN) that creates artistic images of high quality. AffectGAN is driven by a semantic prompt and is able to evoke and express certain emotions. The results obtained through our user study showcase the potential of semantic-based generation of visual art that embeds affective information. Our proof-of-concept experiment illustrates the capacity of CLIP as combined with VQGAN models to generate high resolution images that express particular emotional states or dimensions. The images can be used as emotion elicitors, as training data for augmenting existing affective databases and can even form affective interaction applications for cultural heritage, architecture, visual art and even digital games. AffectGAN also opens up new research avenues at the intersection of affective computing and computational creativity at large, imbuing computational creators with an affective theory of mind \cite{poletti2012cognitive} and allowing them to both generate (through AffectGAN) and assess (e.g. through the ArtEmis classifier \cite{achlioptas2021artemis}) the intended emotions they wish to evoke from the audience. 

\bibliographystyle{IEEEbib}
\bibliography{affect_image}


\end{document}